\documentclass[letterpaper, 10pt, conference]{ieeeconf}
\IEEEoverridecommandlockouts                              
\let\autocite\cite
\usepackage[colorlinks,linkcolor=blue]{hyperref}
\usepackage{graphics} 
\usepackage{epsfig} 
\usepackage{mathptmx} 
\usepackage{times} 
\usepackage{amsmath}
\usepackage{amssymb}
\usepackage{booktabs}
\usepackage{array}
\usepackage{hyperref}
\usepackage{cite}
\usepackage{algorithm}
\usepackage{algpseudocode}

\begin{document}
\title{\LARGE \bf FACET: Fast and Accurate Event-Based Eye Tracking \\Using Ellipse Modeling for Extended Reality}

% \title{\LARGE \bf EPNet: A Lightweight Network for Event-Based Eye Tracking}

\author{Junyuan Ding$^{1}$, Ziteng Wang$^{2}$, Chang Gao$^{3}$, Min Liu$^{2}$, Qinyu Chen$^{4*}$% <-this % stops a space
\thanks{$^{*}$Qinyu Chen is the corresponding author.}
\thanks{$^{1}$Junyuan Ding is with the School of Automation Science and Electrical Engineering, Beihang University, Beijing, China
        {\tt\small dingjunyuan@buaa.edu.cn}}%
\thanks{$^{2}$Ziteng Wang and Min Liu are with DVSense (Beijing) Technology Co., Ltd., China.}%
\thanks{$^{3}$Chang Gao is with the Department of Microelectronics, Delft University of Technology, The Netherlands.}%
\thanks{$^{4}$Qinyu Chen is with the Leiden Institute of Advanced Computer Science (LIACS), Leiden University, The Netherlands.
        {\tt\small q.chen@liacs.leidenuniv.nl}}%
}
\maketitle
\thispagestyle{empty}
\pagestyle{empty}
% \begin{abstract}
% Eye tracking is a key technology in Extended Reality (XR) environments, enabling gaze-based interactions and more immersive experiences. However, XR applications impose strict requirements for power efficiency, low latency, and high accuracy, which traditional CMOS cameras struggle to meet simultaneously. 
% To address these challenges, we propose FAEET, a framework based on event-based cameras that offers fast and accurate pupil detection performance, making it ideal for real-time eye tracking in XR scenarios.
% Current dataset use, We enhance the existing largest event-based eye tracking dataset EV-Eye that uses a semi-supervised approach to expand the annotated data, and converts the mask labels to ellipse-based annotations.

% First, we modified EV-Eye dataset by  that parameterizes the pupil into ellipses and uses a fixed event count combined with a fast causal event volume event processing method to improve accuracy and speed. Secondly, we design a MobileNet-based network and FPN with DSConv structure to ensure the model's lightweight nature and inference speed. Finally, we design a trigonometric loss to address abrupt changes in traditional angle loss and improve accuracy. On the test set with a resolution of 64 $\times$ 64, our method achieves an average distance of 0.2030 pixels between the predicted pupil center and the ground truth, with an average inference time of 0.5302 ms, 37.17\% and 46.68\% higher than segmentation-based methods. The project code is openly available at:~\href{https://github.com/DeanJY/FAEET}{https://github.com/DeanJY/FAEET}
% \end{abstract}
\begin{abstract}
% Eye tracking is a key technology in Extended Reality (XR) environments, enabling gaze-based interactions and more immersive experiences. However, XR applications impose strict requirements for power efficiency, low latency, and high accuracy, which traditional CMOS cameras struggle to meet simultaneously. In event-based eye-tracking methods, segmenting the pupil first , fitting an ellipse and finally tracking is a method that can reduce the number of inferences to achieve high-frequency tracking. FAEET can quickly and accurately output ellipses end-to-end, ensuring accuracy while reducing the time before tracking, making it ideal for real-time eye tracking in XR scenarios.
% We enhance the existing largest event-based eye tracking dataset EV-Eye that uses a semi-supervised approach to expand the annotated data, and converts the mask labels to ellipse-based annotations. Then we design a lightweight network structure to output ellipse directly, with a novel trigonometric loss to address discontinuity in traditional angle prediction and fast causal event volume to reduce event processing time. On the test set with a resolution of 64 $\times$ 64, our method achieves an average distance of 0.2030 pixels between the predicted pupil center and the ground truth, with an average inference time of 0.5302 ms, 37.17\% and 46.68\% higher than segmentation-based methods. The project code is openly available at:~\href{https://github.com/DeanJY/FAEET}{https://github.com/DeanJY/FAEET}

Eye tracking is a key technology for gaze-based interactions in Extended Reality (XR), but traditional frame-based systems struggle to meet XR's demands for high accuracy, low latency, and power efficiency. Event cameras offer a promising alternative due to their high temporal resolution and low power consumption. In this paper, we present FACET (Fast and Accurate Event-based Eye Tracking), an end-to-end neural network that directly outputs pupil ellipse parameters from event data, optimized for real-time XR applications. The ellipse output can be directly used in subsequent ellipse-based pupil trackers. We enhance the EV-Eye dataset by expanding annotated data and converting original mask labels to ellipse-based annotations to train the model. Besides, a novel trigonometric loss is adopted to address angle discontinuities and a fast causal event volume event representation method is put forward. On the enhanced EV-Eye test set,
% a 64$\times$64 resolution test set, 
FACET achieves an average pupil center error of 0.20 pixels and an inference time of 0.53~ms, reducing pixel error and inference time by 1.6$\times$ and 1.8$\times$ compared to the prior art, EV-Eye, with 4.4$\times$ and 11.7$\times$ less parameters and arithmetic operations. The code is available at~\href{https://github.com/DeanJY/FACET}{https://github.com/DeanJY/FACET}.

\end{abstract}

\section{Introduction}
% In recent years, the field of Extended Reality (XR) has witnessed remarkable advancement, revolutionizing the way we interact with our digital surroundings.
% Eye tracking, a technology that measures and records eye movements, has proven to become an indispensable component at the heart of immersive XR experiences, especially since the introduction of the Apple Vision Pro in June 2023~\cite{apple2023visionpro}.
% It detects where a person is looking and how their eyes move across a visual scene, enabling insights into visual attention and cognitive processes. 
% Eye tracking in XR environments enables gaze-based interactions~\cite{fernandes2023leveling,hu2022gaze,adhanom2023eye} allowing users to control and navigate virtual environments by directing their gaze at specific objects or areas of interest.
% Ongoing efforts have been made to implement this technology into wearable devices~\cite{yang2023wearable,menendez2024eye, kurzhals2016visual}. However, issues such as latency and power consumption must be addressed to achieve effective and reliable eye-tracking performance.

Recently, Extended Reality (XR) is rapidly transforming the way people perceive and interact with the digital world. Eye tracking, a technology that measures and records eye movements, has become indispensable for immersive XR experiences~\cite{plopski2022eye,jin2024eye}, especially after the introduction of the Apple Vision Pro in June 2023~\cite{apple2023visionpro}. 
% By detecting where a person is looking and how their eyes move across a scene, it provides insights into visual attention and cognitive processes.
Eye tracking enables gaze-based interactions in XR environments~\cite{fernandes2023leveling,hu2022gaze,adhanom2023eye}, allowing users to control and navigate virtual spaces simply by directing their gaze. While efforts continue to integrate this technology into wearable devices~\cite{yang2023wearable,menendez2024eye,kurzhals2016visual}, challenges like latency and power consumption must be addressed to ensure smooth and effective experience.

% Eye tracking, a critical component of these systems, enables functionalities such as gaze estimation, foveated rendering, varifocal display, and gaze-based user interfaces, significantly enhancing user interaction and immersion~\cite{fernandes2023leveling,fuhl2021teyed}. 
% In wearable medical technology, eye-tracking tasks such as gaze detection and pupil shape detection present novel approaches for both education~\cite{deng2023review} and the monitoring of neurodegenerative diseases like Parkinson's and Alzheimer's~\cite{duan2019dataset,lee2000eye,pretegiani2017eye}.

The human eye is the fastest-moving organ, capable of movements exceeding 300°/s~\cite{verghese2002motion}. Capturing these rapid movements accurately requires a frame rate of kilo-hertz to ensure smooth tracking and reduce motion sickness in virtual environments~\cite{doroudian2023collaboration}. However, achieving such a high frame rate is challenging for wearable devices, which must operate at low power levels, typically in the milliwatt range. 
% Continuous image processing at these rates is computationally intensive and consumes significant power.
Most head-mounted devices (HMDs) rely on frame-based eye-tracking systems. A recent study reports tracking delays between 45 and 81 ms in various HMD eye trackers~\cite{stein2021comparison}, which falls short of the kilo-hertz frame rate needed for accurate eye movement capture. Additionally, frame-based sensors capable of reaching kilo-hertz consume substantial power. The large data volumes also require high bandwidth and significant energy for transfer and processing, posing challenges for real-time applications on wearable devices.

% While XR systems typically employ RGB cameras with fixed frame rates, their performance is affected by deteriorated image quality due to varying light and reflections. Additionally, RGB data are dense and often unsuitable for embedded devices with limited resources, causing inefficient resource utilization, particularly when the pupil is stationary.

Event cameras~\cite{lichtsteiner2008128}, also known as Dynamic Vision Sensors (DVS), offer an effective and efficient alternative for solving eye-tracking challenges. By capturing only brightness changes, they generate sparse asynchronous events, providing high temporal resolution and low power consumption. These unique characteristics make event cameras highly suitable for high-speed, low-power eye tracking: they produce less data and reduce processing needs during fixation while still capturing fast and subtle eye movements during saccades. Previous event-based eye-tracking studies have shown promising results~\cite{chen20233et, bonazzi2024retina, EGaze2024li, angelopoulos2020event, wang2024event, pei2024lightweight, zhao2024ev, li2023track,wang2024mambapupil,stoffregen2022event,lin2024fapnet}. 
% however, they have not fully exploited the inherent properties of event-based data, making it challenging to deliver the required low-power, fast, and accurate eye tracking on mobile devices.
However, most of them use detection neural networks to detect the pupil in every step. The high computational cost of neural networks prevents these models from achieving higher frequency. ~\cite{zhao2024ev,li2023track} use simple ellipse-based trackers to track the pupil for most steps and employ neural network inference only when the pupil tracking is lost, for example after a blink. This detection-tracking schema significantly reduces the computational burden. However, these two methods use a segmentation model to acquire the mask of the pupil and then fit its ellipse boundary. Compared to lightweight detection models like MobileNet series~\cite{howard2017mobilenets,sandler2018mobilenetv2,howard2019searching}, segmentation models (e.g. U-Net~\cite{ronneberger2015u}) have larger computational cost. It also does not take advantage of the fact that event cameras emphasize the boundaries of pupils. 

To fully take advantage of the event data, we propose FACET, \underline{F}ast and \underline{AC}curate Event-based \underline{E}ye \underline{T}racking, a lightweight pupil detector that takes in events and outputs ellipse prediction of pupils, which is not only lighter and faster, but also can be trained end-to-end. This detector can be directly fitted into the existing detection-tracking eye-tracking schema. The main contributions of this work are as follows: 
\begin{itemize} 
\item We introduced a fast and accurate end-to-end event-based pupil detector using ellipse modeling.
\item We proposed a dataset enhancement method that uses a semi-supervised approach to expand the annotated data and convert the mask labels to ellipse-based annotations. We used this method to label all 1.5 million samples in the EV-Eye dataset, while the original dataset only has over 9,000 labeled samples.
\item We proposed trigonometric loss to address the discontinuity problem in angle prediction for ellipse parameters.
\item We designed a fast causal event volume method for event accumulation to regularize the distribution of event representation values.
\end{itemize}
\section{Related Works}
% \subsection{Traditional Eye Tracking Method}
% \subsubsection{EOG-based Eye Tracking Method}
% EOG-based (electrooculography-based) methods measure the electrical potential generated by eye movements. Electrodes are placed around the eyes to detect these potentials. EOG-based methods are less dependent on lighting conditions and can be used in various environments~\cite{gao2010automatic,mannan2016hybrid}. However, they are generally less accurate than video-based methods and can be uncomfortable for users due to the need for electrode placement. 
\subsection{Frame-based Eye Tracking Method}
Traditional frame-based eye tracking typically utilizes frame-based cameras to capture eye movements, with two common approaches: model-based and appearance-based eye tracking. 
Model-based eye tracking~\cite{lai2014hybrid, mestre2018robust,pfeiffer2016model} locates key points corresponding to the eye's geometrical features and fits them to an eye model using optimization techniques. These methods have limitations in headsets, which often require manual calibration and struggle with variations in eye shape and lighting conditions.
The appearance-based method~\cite{mazzeo2021deep,kim2019nvgaze,lee2020deep,zdarsky2021deep,deane2023deep} focuses on the visual appearance of the eye, with a trend of using deep learning techniques to track the eye within the raw image. It requires substantial training data and the model can be computationally intensive and leads to large processing latency. Additionally, these frame-based methods often require high-resolution cameras, which can be both expensive and cumbersome for mobile devices. Moreover, the frame rate of the standard frame-based camera generally peaks at 200 Hz. Cameras with higher frame rates consume significant power, often at watt levels, which exceeds the milli-watt power budget for a mobile eye tracking system.

% A typical setup includes a high-resolution camera and infrared light sources to illuminate the eyes. The captured images are then processed using algorithms to detect and track the position of the pupils~\cite{bonazzi2023tinytracker,kellnhofer2019gaze360,palmero2018recurrent}. Image processing methods are known for their high accuracy and resolution, but often come with high costs, complex setups, and strict lighting conditions, making it challenging to achieve sub-millisecond time resolution on mobile and embedded devices. 

\subsection{Event-based Eye Tracking Method}

Event-based eye tracking utilizes the sparse data stream from DVS for high frame rates with much less bandwidth, offering greater energy efficiency than traditional frame-based systems. 3ET~\cite{chen20233et} introduces a sparse change-based convolutional Long-Short-Term-Memory (LSTM) model for event-based eye tracking, which reduces arithmetic operations by approximately 4.7$\times$, compared to a standard convolutional
LSTM, without losing accuracy, however, uses synthetic event data and fixed time windows, limiting its ability to meet kilo-hertz frame rate demands. 
Retina~\cite{bonazzi2024retina} introduces a neuromorphic approach that integrates a directly trained Spiking Neuron Network (SNN) regression model and leverages a state-of-the-art low power edge neuromorphic processor, achieving 5\,mW power consumption and around 3-pixel error on the INI-30 dataset with 64$\times$64 resolution.
Lightweight models~\cite{pei2024lightweight, wang2024mambapupil} propose spatio-temporal convolution and bidirectional selective recurrent models, respectively, both with approximately 3-pixel error with 60$\times$80 resolution of the 3ET+ dataset~\cite{wang2024event}.
% E-Track~\cite{li2023track} trains a U-Net to classify pupil events and use a region of interest (RoI) mechanism to track the pupil, reducing CNN inference by 96\%, with a 3.68-pixel error.
\cite{stoffregen2022event} tracks the eye movements by detecting and tracking the corneal glint; however, it requires illumination from a flashing light source.

% Hybrid methods combining event and frame data~\cite{angelopoulos2020event, zhao2024ev} achieve high frame rates and satisfactory accuracy but struggle with power consumption.

On the other hand, frameworks~\cite{angelopoulos2020event, zhao2024ev} have combined both frame and event data for eye tracking, utilizing geometric fitting techniques and segmentation networks, respectively. These approaches demonstrate the advantages of combining event data with frame data, achieving both high frame rates and satisfactory accuracy but struggle with addressing the issue of power consumption.

\section{Dataset}
\label{sec:dataset}
\begin{figure}[t]
    \centering
    \includegraphics[width=1\columnwidth]{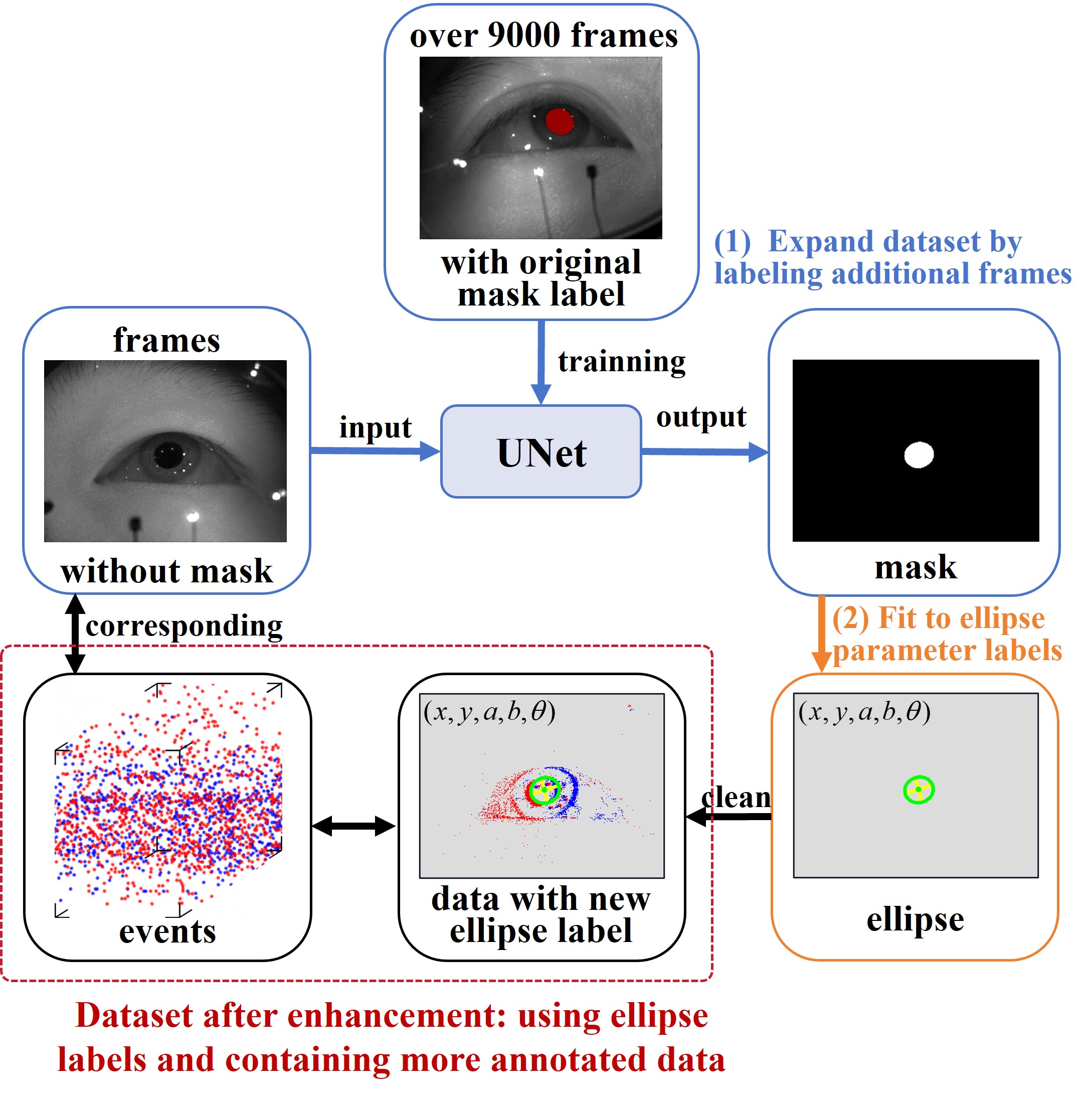}
    \caption{Flowchart for expanding the EV-Eye dataset and annotating it with ellipse labels. We first trained a U-Net segmentation network using over 9,000 frames with mask labels, enabling it to generate masks for other unlabeled frames. Then, we fitted these masks into ellipses to obtain five parameters $(x, y, a, b, \theta)$. Finally, we annotated the events corresponding to these frames with the ellipse labels generated by the U-Net to produce more annotated event data.}
    \label{fig:dataset}
\end{figure}

EV-Eye~\cite{zhao2024ev}, the largest existing event-based eye-tracking dataset, contains data from 48 individuals with a diverse range of genders and ages. The dataset includes over 1.5 million near-eye grayscale images and 2.7 billion event samples captured using two DAVIS346 event cameras. Frames are timestamped at 40 ms intervals, synchronized with the corresponding event data.
Although the EV-Eye dataset~\cite{zhao2024ev} provides a valuable foundation, it has limitations that hinder its direct application to our project. It contains only around 9,000 frames with annotated pupil segmentation, which is insufficient for training robust models. The labels are full-size segmentation masks, while subsequent tracking modules require ellipse predictions as input.

To address these issues, we improved the dataset in two key ways: (1) we expanded it by labeling additional frames, and (2) we converted the full-image pupil segmentation masks into ellipse parameter labels. Fig.~\ref{fig:dataset} illustrates the process of generating the updated dataset.
A semi-supervised learning approach was employed to utilize the large volume of unlabeled data effectively. Pupil segmentations on unlabeled grayscale images were obtained using a U-Net model~\cite{ronneberger2015u} trained on the labeled frames in this dataset.
All pupil segmentation labels are fitted to ellipses expressed in $(x, y, a, b, \theta)$ format, where $(x, y)$ represent the ellipse's center coordinates, $a$ and $b$ are the lengths of the major and minor axes ($a \geq b$), and $\theta \in [0^\circ, 180^\circ)$ denotes the rotation angle. 
% Instances where the pupil area is less than 200 pixels, indicating eye closures or blinks, are discarded. 
Inaccurate labels are manually removed. From the updated EV-Eye dataset, 20,000 samples are randomly selected for the training set, 5,000 for the validation set, and 5,000 for the test set.

% Secondly, Pupil segmentation labels allow for the fitting of an ellipse to determine its parameters: $(x, y, a, b, \theta)$, representing the ellipse's center coordinates $(x, y)$, its major and minor axes $a$ and $b$ (with $a \geq b$), and the rotation angle $\theta \in [-90, 90)$.

% Data cleaning involves the removal of samples with inaccurate labels. Instances where the pupil area is less than 200 pixels, indicative of eye closure or blinks, are discarded. From the EV-Eye dataset, 20,000 samples are randomly selected for the training set， 5,000 for the validation set, and 5000 for the test set.

\begin{figure*}[ht]
    \centering
    \includegraphics[width=1\textwidth]{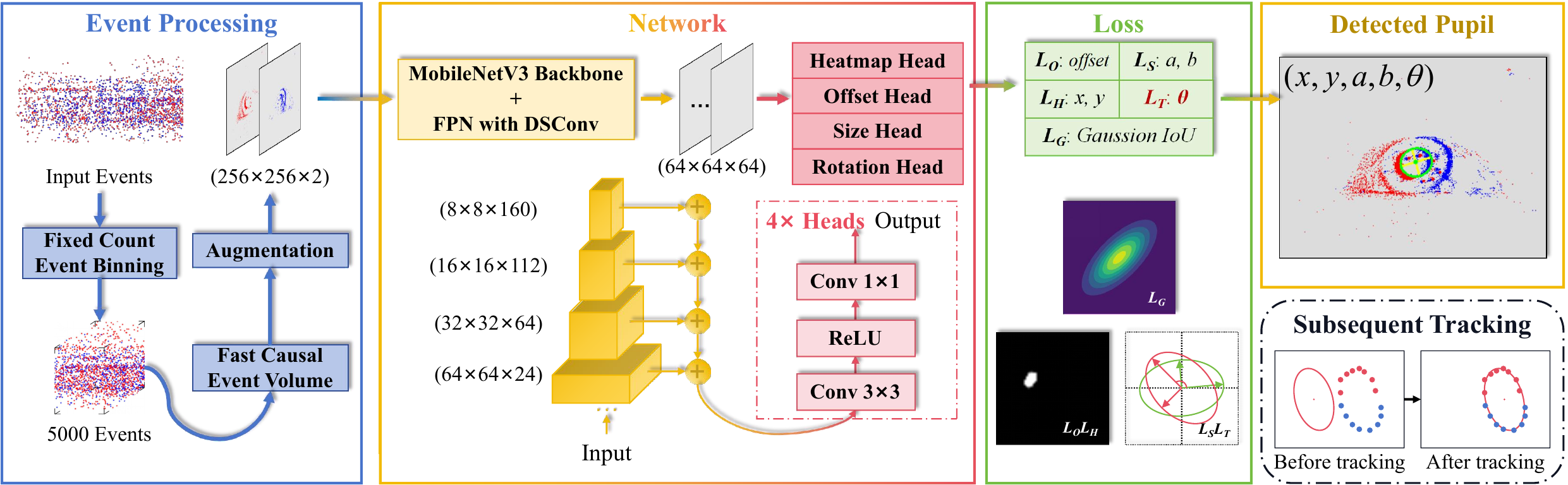}
    \caption{
    % Flowchart of whole FACET. \textbf{Event Processing.} This module includes fixed count binning, fast causal event volume binning, and augmentation; we convert the input events into a frame-like format for subsequent training. \textbf{Network.} We use MobileNetV3 backbone and FPN with DSConv to extract and fuse features, the generated feature map will be sent to the subsequent 4 heads. \textbf{Loss.} The customized loss we designed, trigonometric loss, addresses the discontinuity in traditional angle prediction. Combined with other losses, it effectively measures the difference between the predicted ellipse and the ground truth. \textbf{Detected Pupil.} Unlike segmentation networks that first obtain a mask and then fit an ellipse, FACET is end-to-end and can directly generate ellipses. \textbf{Subsequent Tracking.} Although we did not directly study eye tracking, our direct generation of ellipses provides a foundation for the subsequent use of event-based high-frequency eye-tracking methods\cite{li2023track},\cite{zhaoEVEyeRethinkingHighfrequency2023}. 
    Flowchart of FACET. \textbf{Event Processing:} Input events are converted to a frame-like format using fixed count binning, fast causal event volume, and augmentation for training. \textbf{Network:} A MobileNetV3 backbone with FPN and DSC extracts and fuses features, which are then passed to four heads. \textbf{Loss:} Our total loss function includes several components, among which the customized trigonometric loss $L_T$ plays a crucial role. The term $L_T$ specifically addresses discontinuities in angle prediction, effectively measuring the difference between the predicted ellipse and the ground truth when combined with other losses. \textbf{Detected Pupil:} FACET directly generates ellipses end-to-end, unlike segmentation networks that first obtain a mask and then fit an ellipse. Subsequent Tracking: This direct ellipse generation lays the foundation for high-frequency event-based eye-tracking methods \cite{li2023track,zhaoEVEyeRethinkingHighfrequency2023}.
    }
    \label{fig:overview}
\end{figure*}

\section{Method}
This section covers the processing of events in Section~\ref{sec:event_processing}, the network architecture in Section~\ref{sec:network}, and the loss function in Section~\ref{sec:loss}. An overview of the entire framework is shown in Fig.~\ref{fig:overview}.

\subsection{Event Processing}
\label{sec:event_processing}
\subsubsection{Event Binning Method}

% Events should be divided into bins and accumulated into representations to be used as neural network inputs. 
% Choosing an appropriate binning duration is important, short binning may result in insufficient data, while long binning can reduce the frame rate and introduce too much noise. Therefore, balancing the amount of captured information with the frame rate is essential when binning event data. In \cite{chen20233et, wang2024mambapupil}, fixed time-interval binning is employed to ensure a consistent frame rate. However, when there is no eye movement, no events are generated, yet the model still consumes resources for unnecessary inference. Conversely, during periods of eye movement, a large volume of event data is produced in a short time, placing additional strain on the model.
% In our FACET framework, we utilize a fixed-count binning method. When there is no eye movement, no event data is generated, allowing the model to avoid wasting resources on unnecessary inference. 

To prepare event data for neural network input, events are divided into bins and accumulated into representations. Choosing the appropriate binning duration is important: Short bins may lack sufficient data, while long bins can reduce frame rate and introduce excessive noise. Previous works~\cite{chen20233et, wang2024mambapupil} use fixed time interval binning to maintain consistent frame rates. However, this approach has limitations: when there is no eye movement, no events are generated, yet the model still consumes resources on unnecessary inference; during eye movements, a large volume of events is produced in a short time, increasing computational load.
In our FACET framework, we utilize a fixed-count binning method. This allows the model to avoid wasting resources on unnecessary inference when no events are generated due to the lack of eye movement.

% This makes it more efficient and well-suited for high-frequency tasks like eye movement detection.

% In our FACET, bins are based on a fixed count method. Our experiments have shown that binning with 5000 events provides information content similar to 10000 events in Table\ref{tab:ablation}, achieving similar accuracy with less processing time. Moreover, when there is no eye movement, no event data is generated, and the model does not waste resources on unnecessary inference, making it more suitable for high-frequency inference tasks such as eye movement detection.

\subsubsection{Event Accumulation Method} 
One method to accumulate events in bins into representations is the event volume~\cite{zhu2019unsupervised}. However, the event volume at time $t$ takes events both before and after $t$, which is impossible during real time processing. Due to the temporal causality of event sequences, causal event volume~\cite{pei2024lightweight} using only events before $t$ is more suitable for real-time processing. Building upon this approach, FACET proposes a fast causal event volume that further reduces the time required for event accumulation.

For an event bin $B$ containing n events $E = \{e_i | i = 1\cdots n\}$ in a period of $\Delta t$, where $E_i = (x_i, y_i, p_i, t_i)$ represents the coordinates, polarity, and timestamp of the number $i$ event in the bin, causal event volume accumulates all the events in the bin to a 2D representation. $p_i=0$ means a negative event showing the pixel gets dimmer at that time, while $p_i=1$ means a positive event showing the pixel gets brighter. For the pixel at $(x, y)$, the value of the causal event volume at the end of the event bin $t$ is calculated as follows:
\begin{align}
V_{pos}(x, y)&=\sum_{\{E_i | p_i = 1\}} \delta_{x_i, x} \cdot \delta_{y_i, y} \cdot k\left(\frac{t - t_i}{\Delta t}\right) \\
V_{neg}(x, y)&=\sum_{\{E_i | p_i = 0\}} \delta_{x_i, x} \cdot \delta_{y_i, y} \cdot k\left(\frac{t - t_i}{\Delta t}\right)
\end{align}
$\delta$ represents the Kronecker delta function, in which $\delta_{ij}=1$ only if $i=j$, otherwise $\delta_{ij}=0$. The kernel function $k(\tau)$ and Heaviside step function $H(x)$ are defined as:
\begin{align}
k(\tau) &= H(\tau) \max(1 - |\tau|, 0) \\
H(x) &= 
\begin{cases}
0, & \text{if } x < 0 \\
1, & \text{if } x \geq 0
\end{cases}
\end{align}
here, $V_{pos}$ and $V_{neg}$ represent the accumulated contribution of positive and negative polarity events, respectively, to the occupancy of the bin. 

\begin{figure}[t]
\centering
\includegraphics[width=1\columnwidth]{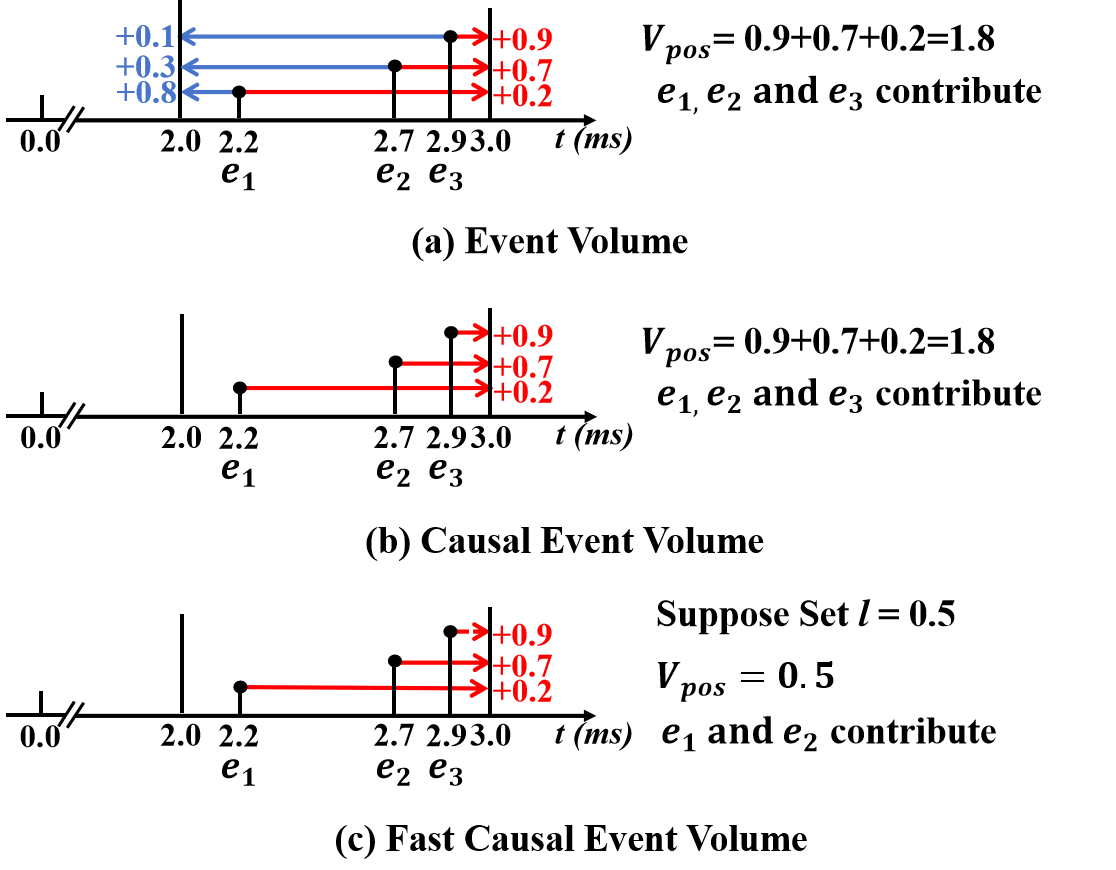}
\caption{Examples of different event accumulation methods: (a) Event Volume, (b) Causal Event Volume, (c) Fast Causal Event Volume. We consider accumulating the events at 3.0~ms timestamp, three events $e_1, e_2, e_3$ with positive polarity occur at 2.2~ms, 2.7~ms and 2.9~ms respectively. Since event volume (a) does not have temporal causality,  these events will also affect the result at 2.0~ms, meaning that future events will influence past time. In the causal event volume example (b), temporal causality is preserved, and all events within the time window are processed.
Our proposed fast causal event volume example (c) introduces a limit $l=0.5$ to optimize the accumulation. This reduces the contribution of earlier events (like $e_1$), speeding up the process for real-time inference, where only $e_1, e_2$ contribute based on the defined limit.}
\label{fig:event_accumulating}
\end{figure}

Accumulating events with setting limits will reduce processing time and benefit for using min-max normalization, contributing to more stable training.
As shown in \textbf{Algorithm 1}, fast causal event volume offers the advantage that once events at the same coordinate reach the defined limit, they are considered to have sufficient information, and no further accumulation is performed, reducing the time to accumulate events. Fig.~\ref{fig:event_accumulating} gives examples of different event accumulation methods: event volume, causal event volume, and fast causal event volume.
\begin{algorithm}[t]
\label{alg}
\caption{Fast Causal Event Volume}
\begin{tabbing}
\textbf{Input:} \hspace{1em} \= $E = \{e_i \ | \ i = 1 \cdots n\}$ ($e_i = (x_i, y_i, p_i, t_i)$) \\
                \> $l$: limit \\
\textbf{Param:} \> $c$: contribution \\
\textbf{Output:} \> $V_{pos}, V_{neg}$ 
\end{tabbing}
\begin{algorithmic}[1]
\For{$e_i$ in $E$}
    \If{$p_i$ is positive}
        \If{$V_{pos}[x_i, y_i] + c_i \leq l$}
            \State $V_{pos}[x_i, y_i] \gets V_{pos}[x_i, y_i] + c_i$
        \EndIf
    \Else
        \If{$V_{neg}[x_i, y_i] + c_i \leq l$}
            \State $V_{neg}[x_i, y_i] \gets V_{neg}[x_i, y_i] + c_i$
        \EndIf
    \EndIf
\EndFor
\State \textbf{return} $V_{pos}, V_{neg}$
\end{algorithmic}
\end{algorithm}

% In FACET, we set $l=255$, the formula is as follows:
% \begin{align}
% V_{F}^+&=\min(l, w \cdot V^+)\\
% V_{F}^-&=\min(l, w \cdot V^-)
% \end{align}

% \subsection{Augmentation}
% In FACET, we employ various data augmentation techniques to improve model generalization, including rotation (R), scaling (S), translation (T), and horizontal flip (HF). These augmentations simulate different distances, angles, positions, and orientations of the eye relative to the event camera. After applying these augmentations, the input’s width (w) and height (h) are uniformly resized to 256 × 256, preparing it for use as input in the subsequent network.
% Table~\ref{tab:data_augmentation} summarizes the data augmentation methods, together with their respective parameters and probabilities.
% \begin{table}[h]
%     \centering
%     \caption{Augmentation Methods}
%     \label{tab:data_augmentation}
%     \begin{tabular}{ccc}
%     \toprule
%     \textbf{Augmentation} & \textbf{Parameters} & \textbf{Probabilities} \\
%     \midrule
%     Rotation & [-15°, 15°] & 1 \\
%     Scaling & 20\%  & 1 \\
%     Translation & 20\%  & 1 \\
%     Horizontal Flip & None & 0.5 \\
%     \bottomrule
%     \end{tabular}
% \end{table}
\subsection{Network}
\label{sec:network}
The network is designed with a focus on lightweight architecture. We use MobileNetV3~\cite{howard2019searching} as the backbone for feature extraction, taking advantage of depthwise separable convolution (DSC) blocks to reduce complexity.
Furthermore, we accelerate the Feature Pyramid Network (FPN)~\cite{lin2017feature} by replacing traditional convolution blocks with DSC blocks, enhancing overall performance.

% \subsubsection{DSC Block}
% DSC blocks reduce computational cost and model size while maintaining performance. They consist of depthwise convolution (DWC) followed by pointwise convolution (PWC). DWC applies convolution independently to each input channel, while PWC is a $1 \times 1$ convolution that combines the outputs of DWC across different channels. DSConv can be represented as \eqref{eq:dsconv}:
% \begin{equation}
% \begin{aligned}
% Z &= \text{DSC}(X, F_{dw}, F_{pw}) \\
%   &= \text{PWC}(\text{DWConv}(X, F_{dw}), F_{pw})
% \end{aligned}
% \label{eq:dsconv}
% \end{equation}

% \noindent where $Z$ is the final output feature map.
% In conclusion, DSC reduces computational complexity from $O(HWC^2)$ to $O(HWC + C^2)$, enhancing feature extraction and integration efficiency. This optimization improves model efficiency and makes deployment in resource-constrained environments more feasible such as mobile and embedded systems.

\subsubsection{MobileNetV3 Backbone}
MobileNetV3 reduces model size using DSC and enhances model expressiveness with squeeze-and-excitation (SE) blocks, achieving high inference efficiency and high. MobileNetV3 has been proven to be an excellent backbone for models working on edge devices like mobile phones and XR devices. 

% Integrated with FPN, MobileNetV3 leverages feature maps at different resolutions to improve multi-scale object detection.

% The SE block re-calibrates channel-level features by modeling interdependence between channels. The SE block can be represented as \eqref{eq:se}:

% \begin{equation}
% \begin{aligned}
% \mathbf{z} &= \text{GlobalAvgPool}(\mathbf{X}) \\
% \mathbf{s} &= \sigma(\mathbf{W}_2 \delta(\mathbf{W}_1 \mathbf{z})) \\
% \mathbf{X'} &= \mathbf{X} \cdot \mathbf{s}
% \end{aligned}
% \label{eq:se}
% \end{equation}
% where \(\mathbf{X}\) is the input feature map, \(\mathbf{z}\) is the channel-level global average pooling result, \(\mathbf{W}_1\) and \(\mathbf{W}_2\) are weight matrices, \(\delta\) is the ReLU activation, \(\sigma\) is the Sigmoid activation, and \(\mathbf{s}\) is the channel-level scaling factor. The recalibrated feature map \(\mathbf{X'}\) is obtained by multiplying \(\mathbf{X}\) with \(\mathbf{s}\). SE blocks improve model performance while keeping the model lightweight.

\subsubsection{FPN with DSC}
We replace all normal convolution blocks in FPN with DSC, reducing the parameters of FPN and improving the speed of feature fusion. The FPN of FACET can be represented as \eqref{eq:fpn}:
\begin{equation}
\label{eq:fpn}
P_i = \text{DSC}(C_i) + \text{Upsample}(P_{i+1}), \quad  i\in\{5,4,3,2\} \\
\end{equation}

\noindent where \(C_i\) represents the feature maps from different stages of MobileNetV3, and \(P_i\) represents the corresponding feature maps in the FPN. 

DSC blocks consist of depthwise convolution (DWC) followed by a $1\times1$ convolution, extracting features space-wise and channel-wise respectively. DSC reduces computational complexity from $O(HWC^2)$ to $O(HWC + C^2)$.
The final feature map $P_2$ with $(64, 64, 64)$ dimension is used in procedures afterward.

\subsubsection{Heads}
The output feature map of the FPN is processed by four detection heads: heatmap head, offset head, size head and rotation head. Each head is composed of a 3x3 convolution, ReLU, and a 1$\times$1 convolution, as shown in fig.~\ref{fig:overview}. 
The heatmap head outputs a $64\times64$ heatmap which is used to predict the center of the pupil. The location with the maximum value in the heatmap is considered to be the center of the ellipse. Then an offset of the center is predicted by the offset head to refine the ellipse center, compensating for the quantization error from the limited resolution. The size head predicts the major and minor axis $(a, b)$ of the ellipse. The rotation head predicts the rotation of the ellipse. Raw rotation predictions are presented in the form of $\hat{\vec{r}} = (\hat{\sin{(2\theta)}}, \hat{\cos{(2\theta)}})$. Then we normalize the prediction $\vec{r}=\hat{\vec{r}}/||\hat{\vec{r}}||_2$ and use $\vec{r}$ to recover the rotation $\theta$ of the ellipse. The reason we choose this format of rotation representation is explained in \ref{Subsec: Trigonomettric Loss}.
% In conclusion, the network design of E3T makes it more feasible and performant for deployment in resource-constrained environments, such as mobile devices and embedded systems.

\subsection{Loss}
\label{sec:loss}
In FACET, we designed a comprehensive loss function to enhance the model performance, defined as follows:
\begin{equation}
\label{eq:loss}
L = \lambda_H L_H + \lambda_O L_O + \lambda_S L_S + \lambda_G L_G + \lambda_T L_T
\end{equation}
where $L_H$ is the Heatmap Loss, $L_O$ is the Offset Loss, $L_S$ is the Size Loss, and $L_G$ is the Gaussian IoU Loss, and $L_T$ is the Trigonometric Loss. 
The innovative aspect of our approach is the introduction of Trigonometric Loss ($L_T$), which significantly improves angle prediction by addressing the discontinuity in traditional angle loss computation. 
% This enhances training stability and results in more accurate ellipse angle predictions.

% We designed a new loss function for angle prediction: Trigonometric Loss. Traditional loss functions directly compute the L1 loss for angles, which is simple and straightforward. However, using either degrees or radians leads to discontinuities, causing instability during training. E3T introduces a new loss function called Trigonometric Loss, which mitigates the discontinuities in angle loss computation and improves the accuracy of angle predictions.
\subsubsection{Trigonometric Loss}
\label{Subsec: Trigonomettric Loss}
We define the rotation of an ellipse as placing the major axis of the ellipse horizontally and rotating the ellipse around its center at an angle $\theta \in [0^\circ, 180^\circ)$. Due to the symmetry of ellipses and the periodicity of rotations, $0^\circ$ and $180^\circ$ represent the same ellipse, which means the two ends of the range $[0, 180)$ should be continuous. Regular loss functions usually measure the norm of the difference between the prediction and the ground truth, leading to a huge discontinuity at the two ends. This discontinuity results in a large gradient event if the real difference between the prediction and the ground truth is small. The mismatch harms the training of the model. 

To deal with this discontinuity, we propose Trigonometric Loss. The model predicts $\hat{\vec{r}}_p=(\hat{\sin{(2\theta)}}, \hat{\cos{(2\theta)}})$. The trigonometric loss $L_T$ calculates the L2 loss between $\hat{\vec{r}}_p$ and the ground truth $\vec{r}_g$
% \eqref{eq:tri}
:
\begin{align}
L_T &= L_2(\hat{\vec{r}}_p, \vec{r}_g)
\label{eq:tri}
\end{align}
% and traditional Angle Loss is \eqref{eq:angle}, where the units of $\theta_p$ and $\theta_g$ are radians.
% \begin{equation}
% L_A = \text{smoothL1}(\theta_p, \theta_g)
% \label{eq:angle}
% \end{equation}

\begin{figure}[t]
    \centering
    \includegraphics[width=0.9\columnwidth]{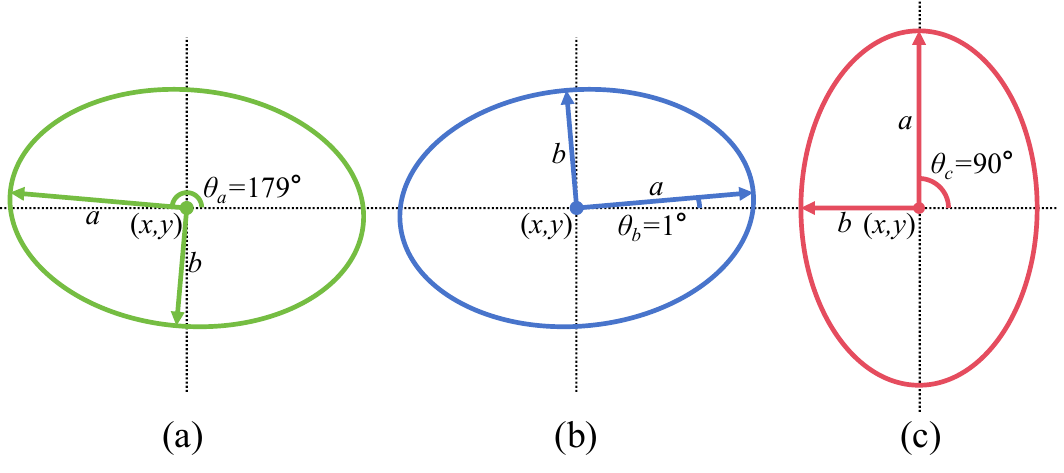}
    \caption{Examples of ellipses at different angles. (a) $\theta_a=179^\circ$, (b) $\theta_b=1^\circ$, and (c) $\theta_c=90^\circ$. 
    Although $179^\circ$ and $1^\circ$ differ numerically, they produce ellipses more similar to each other than to $90^\circ$, implying their corresponding loss should reflect this pattern.}
    \label{fig:ellipse}
\end{figure}

This mapping from $\theta$ to $(\sin{(2\theta)}, \cos{(2\theta)})$ transfers the discontinuous domain $[0, 180)$ to a continuous 2D domain. For example, in Fig.~\ref{fig:ellipse}, $\theta_a=179^\circ$ should be similar to $\theta_b=1^\circ$. Therefore, the loss should be small. But $\theta_c=90^\circ$ should be very different from $\theta_a$ and $\theta_b$, thus should have a big loss.
As is shown in Table~\ref{tab:comparison}, if we compare $L_T$ with regular L1 angle loss $L_A = L_1(\theta_p, \theta_g)$, we can see that $L_A(\theta_a, \theta_b)$ is even bigger than $L_A(\theta_a, \theta_c)$ and $L_A(\theta_b, \theta_c)$. In contrast, $L_T$ offers a more reasonable loss, where $L_T(\theta_a, \theta_b) \simeq 0$ and $L_T(\theta_a, \theta_c) = L_T(\theta_b, \theta_c) \gg L_T(\theta_a, \theta_b)$.
\begin{table}[t]
    \centering
    \caption{Comparison of $L_T$ and $L_A$ between different angles}
    \label{tab:comparison}
    \begin{tabular}{l l l l}
    \toprule
    & \textbf{$\theta_a, \theta_b$} & \textbf{$\theta_a, \theta_c$} & \textbf{$\theta_b, \theta_c$} \\
    \midrule
    $L_A$ & 3.1067 & 1.5533 & 1.5533 \\
    $L_T$ & 0.0049 & 3.9988 & 3.9988 \\
    \bottomrule
    \end{tabular}
\end{table}

\subsubsection{Other Loss Components}
Beyond the proposed Trigonometric Loss, the total loss function incorporates the Heatmap Loss $L_H$, Offset Loss $L_O$, and Size Loss $L_S$ from CenterNet~\cite{duan2019centernet} and Gaussian IoU Loss $L_G$ from ElDet~\cite{wangElDetAnchorfreeGeneral2022}.
$L_H$ is the focal loss of the heatmap. $L_O$ and $L_S$ are smooth L1 losses between the predicted offset, scale and their corresponding ground truth. Gaussian IOU loss $L_G$ is proposed in ElDet. An ellipse bounding box $B(x, y, a, b, \theta)$ can be reformulated in a 2D Gaussian distribution $G(\mu, \Sigma)$. $L_G$ is measured using the wasserstein distance~\cite{panaretos2019statistical} between the predicted distribution and the ground truth distribution. 
 
\section{Experimental Results}
\begin{table*}[t]
    \centering
    \caption{Comparison of accuracy, parameters, GFLOPs, and inference time. The best metric is in \textbf{bold}, and the second best is \underline{underlined}.}
    \label{tab:results}
    \begin{tabular}{l c c c c r r r}
    \toprule
    \textbf{Method} & $\mathbf{P_{10}}$ (\%) & $\mathbf{P_{5}}$ (\%) & $\mathbf{P_{1}}$ (\%) & $\mathbf{PE}$ (pixel)& \textbf{Params} & \textbf{GFLOPs} & \textbf{Inf. Time} (ms)\\
    \midrule
    E-Track & 99.17 & 98.28 & 79.22 & 1.6680 & 17.27 M & 40.19 & 0.9443 \\
    EV-Eye & 99.92 & 99.91 & 98.87 & 0.3231 & 17.27 M  & 40.11 & 0.9438 \\
    ElDet & 99.76 & 99.48 & 95.32 & 0.6273 & 16.82 M & 8.30 & \textsuperscript{*}12.3854 \\
    TennSt & 98.55 & 96.77 & 73.67 & 1.1291 & \textbf{0.81 M} & \underline{5.49} & \textbf{0.3384} \\
    FACET (ours) & \textbf{100} & \textbf{99.98} & \textbf{99.59} & \textbf{0.2030} & \underline{3.92 M} & \textbf{3.44} & \underline{0.5302} \\
    \bottomrule
    \end{tabular}
    \flushleft\footnotesize{\textsuperscript{*}The inference time for ElDet is obtained using PyTorch, as ElDet includes a custom module that is incompatible with official TensorRT.}
\end{table*}
\begin{figure}[t]
    \centering
    \includegraphics[width=0.9\columnwidth]{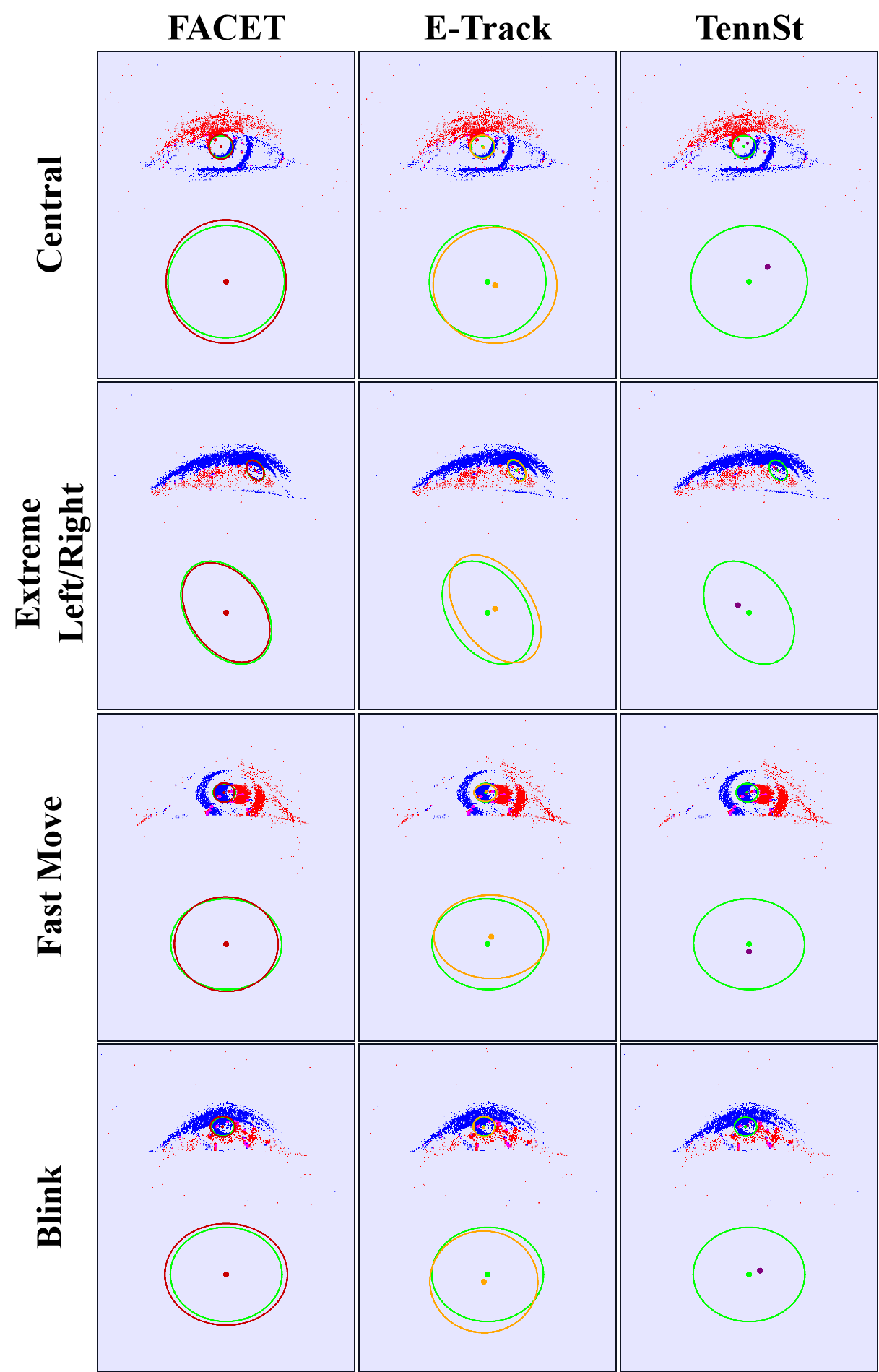}
    \caption{Visual comparison of E-Track, TennSt, and our FACET in four typical scenarios.}
    \label{fig:results}
\end{figure}
\begin{table}[t]
    \centering
    \caption{Ablation study Results}
    \label{tab:ablation}
    \begin{tabular}{l l l l }
    \toprule
     & $\mathbf{P_{1}}$ (\%) & $\mathbf{PE}$ (pixels) & $\mathbf{EPT}$ (ms) \\
    \midrule
    \textbf{FACET (Ours)} & 99.59 & 0.2030 & 1.6493  \\
    \\
    \multicolumn{4}{l}{\textbf{Event Accumulating Method} - Fast causal event volume} \\
    -Causal event volume & 99.59 & 0.2193 & 1.7799 \\
    -Event volume & 98.81 & 0.2500 & 1.8502  \\
    \\
    \multicolumn{4}{l}{\textbf{Event Binning Method} - Fixed Count 5000\phantom{0}evts} \\
    -500\phantom{000}evts & 93.52 & 0.4326 & 0.8311  \\
    -1000\phantom{00}evts & 96.43 & 0.3147 & 1.0064  \\
    -2000\phantom{00}evts & 98.43 & 0.2453 & 1.2894  \\
    -10000\phantom{0}evts & 99.89 & 0.2194 & 2.8983  \\
    -10000\phantom{0}$\mu \text{s}$ & 97.29 & 0.2886 & 0.8480  \\
    \\
    % \multicolumn{4}{l}{\textbf{FPN} - FPN with DSConv} \\
    % -Normal Conv2d & 99.60 & 0.2330 & \multicolumn{1}{c}{-} \\
    % \\
    \multicolumn{4}{l}{\textbf{Loss} - Trigonometric Loss} \\
    -Angle Loss & 98.90 & 0.2878 & \multicolumn{1}{c}{-} \\
    % \\
    % \multicolumn{4}{l}{\textbf{Augmentation} - All} \\
    % -w/o Rotation & 0.9943 & 0.2326 & \multicolumn{1}{c}{-} \\
    % -w/o Scale & 0.9939 & 0.2206 & \multicolumn{1}{c}{-} \\
    % -w/o Translation & 0.9954 & 0.2271 & \multicolumn{1}{c}{-} \\
    % -w/o Horizontal Flip & 0.9952 & 0.2083 & \multicolumn{1}{c}{-} \\
    % -w/o All & 0.9770 & 0.3464 & \multicolumn{1}{c}{-} \\
    \bottomrule
    \end{tabular}
\end{table}
% We evaluate the performance of FACET through a series of experiments. All metrics presented represent the average values obtained over 10 repetitions. In the tables, bolded values indicate the best performance, while underlined values represent the second-best results.
We evaluate our model and other comparative models: E-track~\cite{li2023track}, EV-Eye~\cite{zhao2024ev}, ElDet\cite{wangElDetAnchorfreeGeneral2022}, and TennSt~\cite{pei2024lightweight}, on the enhanced EV-Eye dataset described in Section~\ref{sec:dataset}. All metrics are obtained at a resolution of 64 $\times$ 64.

\subsection{Training Details}
We train our models implemented with PyTorch \cite{paszke2019pytorch} on a single NVIDIA RTX 3090 GPU.
We used a batch size of 32, with 70 training epochs. The optimizer is Adam, with an initial learning rate of $1 \times 10^{-3}$ and a weight decay of $1 \times 10^{-5}$. For the first five epochs, the warm-up learning rate is $1 \times 10^{-5}$, and the learning rate will decay by a factor of 0.7 every 10 epochs thereafter. 
For the proposed fast causal event volume, we set the limit $l$ to be 25.
We apply data augmentation techniques such as rotation, scaling, translation, and horizontal flip to simulate varying distances, angles, positions, and orientations of the eye relative to the event camera, improving model generalization.

% We also apply various data augmentation techniques to improve model generalization, including rotation (R), scaling (S), translation (T), and horizontal flip (HF). These augmentations simulate different distances, angles, positions, and orientations of the eye relative to the event camera.
% Table~\ref{tab:data_augmentation} summarizes the data augmentation methods, together with their respective parameters and probabilities.
% \begin{table}[h]
%     \centering
%     \caption{Augmentation Methods}
%     \label{tab:data_augmentation}
%     \begin{tabular}{ccc}
%     \toprule
%     \textbf{Augmentation} & \textbf{Parameters} & \textbf{Probabilities} \\
%     \midrule
%     Rotation & [-15°, 15°] & 1 \\
%     Scaling & 20\%  & 1 \\
%     Translation & 20\%  & 1 \\
%     Horizontal Flip & None & 0.5 \\
%     \bottomrule
%     \end{tabular}
% \end{table}

\subsection{Accuracy Results}
In Table~\ref{tab:results}, $P_{n}$ ($n \in \{10, 5, 1\}$) represents the probability that the predicted pupil center is within $n$ pixels of the true center, and Pixel Error (PE) represents the average distance from the predicted pupil center to the true center, measured in pixels. 
It can be seen that FACET achieved the best performance in all metrics, with a 0.2030-pixel error, far surpassing other methods. 
% This indicates that FACET has a significant advantage in the precision of pupil center prediction.
This is reflected in the visualized results in Fig.~\ref{fig:results}. We selected four typical scenarios: central, extreme right/left, fast move, and blink, and compared FACET with E-Track and TennSt. The visual results proved that the FACET achieves the highest accuracy.
% \begin{table}[t]
%     \centering
%     \caption{Comparision of model accuracy. The best metric is in \textbf{bold}.}
%     \label{tab:comparision_results}
%     \begin{tabular}{l l l l l}
%     \toprule
%     \textbf{Method} & \textbf{$P_{10}$} & \textbf{$P_{5}$} & \textbf{$P_{1}$} & PE \\
%     \midrule
%     E-Track & 0.9917 & 0.9828 & 0.7922 & 1.6680 \\
%     EV-Eye &0.9992 & 0.9991 & 0.9887 & 0.3231 \\
%     ElDet & 0.9976 & 0.9948 & 0.9532 & 0.6273 \\
%     TennSt & 0.9855 & 0.9677 & 0.7367 & 1.1291 \\
%     FACET(ours) & \textbf{1} & \textbf{0.9998} & \textbf{0.9959} & \textbf{0.2030} \\
%     \bottomrule
%     \end{tabular}
% \end{table}

    % \begin{table}[t]
%     \centering
%     \caption{Comparision of model parameters, GFLOPs and inference time. The best metric is in \textbf{bold}, the second best is \underline{underlined}.}
%     \label{tab:comparision_size}
%     \begin{tabular}{l r r r}
%     \toprule
%     \textbf{Method} & \textbf{Params} & \textbf{GFLOPs} & \textbf{Inf. Time (ms)}\\
%     \midrule
%     E-Track & 17.27 M & 40.19 & 0.9443 \\
%     EV-Eye & 17.27 M  & 40.11 & 0.9438 \\
%     ElDet & 16.82 M & 8.30 & \textsuperscript{*}12.3854 \\
%     TennSt & \textbf{0.81 M} & \underline{5.49} & \textbf{0.3384} \\
%     FACET(ours) & \underline{3.92 M} & \textbf{3.44} & \underline{0.5302} \\
%     \bottomrule
%     \end{tabular}  
%     \flushleft\footnotesize{\textsuperscript{*}The result is obtained using PyTorch, because ElDet includes a custom module that is not compatible with official TensorRT.}
% \end{table}

\subsection{Efficiency Results}
Efficiency is evaluated using three metrics: model parameters, number of operations and inference time, with the latter measured per sample, accelerated by TensorRT on an RTX 3090.
As shown in Table~\ref{tab:results}, 
FACET outperforms other models in both the number of operations (3.44\,GFLOPs) and the inference time (0.5302\,ms), while also having the second smallest parameter count (3.92\,M).
E-Track and EV-Eye have similar parameter counts, GFLOPs, and inference times. Compared to FACET, they require 4.4$\times$ and 11.7$\times$ more parameters and arithmetic operations, respectively, and their inference time is 1.8$\times$ longer than that of FACET.
ElDet has 4.3$\times$ more parameters and 2.4$\times$ more GFLOPs than FACET.
TennSt, a fully convolutional network, has the lowest parameter count (0.81\,M) and the fastest inference time (0.3384\,ms). However, it requires 5.49\,GFLOPs, 1.6$\times$ more than FACET, and is incompatible with the subsequent ellipse-based tracking module. Furthermore, TennSt has suboptimal accuracy, achieving a $P_1$ score of 73.67\%.

\subsection{Ablation Studies}

Table~\ref{tab:ablation} presents the results of the ablation study. EPT denotes the event processing time in milliseconds. FACET performs well across all metrics, achieving a $P_1$ of 99.59\%, a PE of 0.2030 pixels, and an EPT of 1.6493 ms.
For event accumulation, we found that traditional event volume and causal event volume methods increased the EPT by 0.20~ms and 0.13~ms, respectively. In contrast, using fast causal event volume leads to a decrease in all metrics.
For the event binning method, fixing the event count to 5000 provides the best-balanced performance. Reducing the count to 500 reduces $P_1$ to 93.52\% and increases PE to 0.4326 pixels, indicating that fewer events do not capture enough information. Increasing the count to 10,000 raises $P_1$ to 99.89\%, but also increases the EPT to 2.8983 ms, nearly 1.8$\times$ longer than using 5000 events. Using a fixed time interval reduces the EPT to 0.8480 ms, but with variable event counts, the accuracy drops to $P_1$ of 97.29\%. 
% In the FPN structure, replacing DSConv with Normal Conv2d slightly improves $P_{1}$ to 99.60\% but increases PE to 0.2330 pixels. 
Using Angle Loss instead of Trigonometric Loss lowers $P_1$ to 98.90\% and increases PE to 0.2878 pixels, confirming that our loss design improves accuracy.

% In the Augmentation method, R, S, T and HL significantly improved the model's performance.

% Our experiments demonstrate that binning with 5,000 events provides comparable information to 10,000 events, achieving similar accuracy while reducing processing time.

% In summary, if the goal is to achieve maximum detection speed, one can use a smaller number of events, such as 2000 events. Conversely, if the priority is to enhance detection accuracy, a higher number of events can be used, such as 10000 events, and the convolutional blocks within the FPN can be substituted with standard 2D convolutional blocks from DSConv.

% \subsection{Visualization of Results}
% \begin{figure}[h]
%     \centering
%     \includegraphics[width=1\columnwidth]{figures/results_new.png}
%     \caption{Visualization of Results}
%     \label{fig:results}
% \end{figure}

% In Figure \ref{fig:results}, the upper half of each image shows the detection results at the original resolution, while the lower half presents a proportional enlargement of the detected areas. The green ellipses and their center points represent the ground truth, the red ellipses and their center points denote the predicted outcomes of FACET, and the yellow points indicate the center point of the ellipses predicted by TennSt. It is evident that FACET effectively fits the pupil boundary and demonstrates greater accuracy and stability in predicting the center point compared to TennSt.
\section{Conclusion}
This work enhances the existing event-based eye-tracking dataset EV-Eye and proposes a fast and accurate eye-tracking solution: FACET, using pure event data. FACET directly outputs ellipses accurately and quickly for subsequent tracking. It uses fast causal event volume to reduce event processing time and a novel trigonometric loss to address the discontinuity in traditional angle prediction.
Our experiments demonstrate that FACET is competitive in efficiency while achieving superior accuracy among the state-of-the-art methods, which highlights FACET's significant potential for eye tracking in XR environments.
In future work, we aim to integrate FACET into an optimized XR system using neural processing units and event-based sensors, enabling seamless real-time eye tracking on headsets.
\section{Acknowledgment}
Thank you to Weining Ren for the valuable suggestions provided for this article.

\bibliographystyle{IEEEtran}
\bibliography{E3T.bib}

\end{document}